\documentclass[12pt]{article}

\usepackage{arxiv}

\usepackage[utf8]{inputenc}
\usepackage[T1]{fontenc}
\usepackage{hyperref}
\usepackage{url}
\usepackage{booktabs}
\usepackage{amsfonts}
\usepackage{graphicx}
\usepackage{subfigure}
\usepackage{multirow}
\usepackage[numbers]{natbib}
\usepackage{enumitem}
\usepackage{array}

\newcolumntype{L}[1]{>{\raggedright\let\newline\\\arraybackslash\hspace{0pt}}m{#1}}
\newcolumntype{C}[1]{>{\centering\let\newline\\\arraybackslash\hspace{0pt}}m{#1}}
\newcolumntype{R}[1]{>{\raggedleft\let\newline\\\arraybackslash\hspace{0pt}}m{#1}}

\newcommand{\eqref}[1]{(\ref*{#1})}

\title{C-XGBoost: A tree boosting model for causal effect estimation}

\date{}                    

\author{ 		
	\href{https://orcid.org/0000-0003-1729-4124}{\includegraphics[scale=0.06]{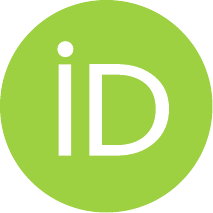}\hspace{1mm}Niki Kiriakidou}\\
	Department of Informatics and Telematics,\\
	Harokopio University of Athens,\\
	Athens, GR 177-78.\\
	\texttt{kiriakidou@hua.gr} \\  	
	\And
	\href{https://orcid.org/0000-0002-3996-3301}{\includegraphics[scale=0.06]{orcid.pdf}\hspace{1mm}Ioannis E.~Livieris}\thanks{Corresponding author}\\
    Department of Statistics and Insurance,\\
    University of Piraeus,\\
    Piraeus, GR 185-34.\\
    \texttt{livieris@unipi.gr} \\
    \And
	\href{https://orcid.org/0000-0002-2461-1928}{\includegraphics[scale=0.06]{orcid.pdf}\hspace{1mm}Christos Diou}\\
	Department of Informatics and Telematics,\\
	Harokopio University of Athens,\\
	Athens, GR 177-78.\\
	\texttt{cdiou@hua.gr} \\  			
}

\renewcommand{\shorttitle}{C-XGBoost: A tree boosting model for causal effect estimation}

\hypersetup{
    pdftitle={C-XGBoost: A tree boosting model for causal effect estimation},
    pdfsubject={cs.CV, cs.LG, cs.AI},
    pdfauthor={Niki Kiriakidou, Ioannis E. Livieris and Christos Diou},
    pdfkeywords={Causal inference, XGBoost, treatment effect estimation, potential outcomes},
}

\begin{document}
    \maketitle

\begin{abstract}
	Causal effect estimation aims at estimating the Average Treatment Effect as well as the Conditional Average Treatment Effect of a treatment to an outcome from the available data. This knowledge is important in many safety-critical domains, where it often needs to be extracted from observational data. In this work, we propose a new causal inference model, named C-XGBoost, for the prediction of potential outcomes. The motivation of our approach is to exploit the superiority of tree-based models for handling tabular data together with the notable property of causal inference neural network-based models to learn representations that are useful for estimating the outcome for both the treatment and non-treatment cases. The proposed model also inherits the considerable advantages of XGBoost model such as efficiently handling features with missing values requiring minimum preprocessing effort, as well as it is equipped with regularization techniques to avoid overfitting/bias. Furthermore, we propose a new loss function for efficiently training the proposed causal inference model. The experimental analysis, which is based on the performance profiles of Dolan and Mor{\'e} as well as on post-hoc and non-parametric statistical tests, provide strong evidence about the effectiveness of the proposed approach. \\ \\
	*** This paper has been accepted for presentation at \textit{IFIP International Conference on Artificial Intelligence Applications and Innovations}. Cite: Kiriakidou, N., Livieris I.E. \& Diou, Ch. (2024). C-XGBoost: A tree boosting model for causal effect estimation. \textit{IFIP International Conference on Artificial Intelligence Applications and Innovations}.***
\end{abstract}


\section{Introduction}

Over the past decade, the growing availability of large datasets and the growing advances in artificial intelligence highlighted the estimation of causal effect as a fundamental research objective.
The primary goal of causal effect estimation (also known as treatment effect estimation) is the identification of the effect of an intervention, which is often called ``\textit{treatment}'' to an outcome. The evaluation process of intervention decisions constitutes a major challenge in many diverse fields, from economics and medicine to education and engineering \cite{johansson2022generalization}. This process requires the study of the difference between alternative choices of intervention, which is not possible to study directly,  since the only available observed outcome is the one of the action actually taken, while the rest remain unknown (counterfactual outcomes). 

Therefore, the prediction of the potential outcomes of unexplored interventions are considered essential for the successful estimation of the causal effects \cite{rubin2005causal}. 
Quite frequently, the outcome is influenced by a variety of factors inclusive of confounders, mediators as well as the treatment itself; hence, rendering its accurate estimation an even more complex task in case of the absence of meticulous adjustment of the identified factors \cite{pearl2009causality}. As a result, the ML models dedicated for estimating causal effects must deal with issues related to confounding factors and sampling biases, which are usually present in these data. Therefore, these challenges underscore the significance of developing sophisticated methodologies and advanced techniques to exploit and leverage observational data for extracting meaningful insights from them.

The rise and advances of deep learning revolutionized the development of prediction models in many scientific fields \cite{pouyanfar2018survey,shinde2018review}. 
Along this line, a variety of neural network-based models such as Dragonnet \cite{shi2019adapting}, TARnet \cite{shalit2017estimating} and NEDnet \cite{shi2019adapting} as well as their modifications \cite{kiriakidou2022improved,kiriakidou2023integrating} have been proposed for estimating causal effects, providing promising performance. A considerable advantage of these models is the lack of requirement of any information about the treatment value for the estimation of the potential outcomes due to their sophisticated architecture design. Therefore, they are able to implicitly handle confounding variables and capture their influence on the outcomes without requiring explicit adjustment. As a result, they are less sensitive to misspecification, as they focus solely on capturing the underlying relationships and patterns within the data, rather than explicitly accounting for the treatment.

Although, deep learning has enabled significant progress, its superiority over tree-based models on tabular data is under question for the following reasons:
Firstly, neural network-based models are usually biased to overly smooth solutions and toward low-frequency functions, while decision tree-based models are better at handling irregular patterns in the data (non-smooth target functions) since they acquire knowledge of piece-wise constant functions \cite{grinsztajn2022tree}; 
and secondly, neural networks are considerably affected by uninformative features in contrast to tree-based models \cite{atashgahi2023supervised}. 
In addition, a research evaluation study \cite{kiriakidou2022evaluation} compared the performance of the most efficient neural network-based and tree-based causal inference models of several collections of benchmarks and highlighted that in many cases the latter exhibited superior performance. By taking these into consideration, we are able to conclude that a new approach focusing on exploiting the advantages of both tree-based and neural network-based approaches may leed to the development of powerful causal inference models.

In this work, motivated by the superiority of tree-based models for handling tabular-based problems, we proposed a new model, named Causal eXtreme Gradient Boosting (C-XGBoost), for causal effect estimation. The main idea behind the proposed approach is to exploit the strong prediction abilities of XGBoost algorithm together with the remarkable property to learn representations that are useful for estimating the outcome for both the treatment and non-treatment cases. The advantages of the proposed causal inference model is that it is able to efficiently handle features with missing values requiring minimum processing effort as well as it is equipped with regularization techniques to avoid overfitting/bias. Furthermore, a new loss function is proposed in order to train the C-XGBoost model. The proposed causal inference model was evaluated against the most efficient and widely used tree-based and neural network-based models on two popular causal inference collections of datasets. The presented experimental analysis demonstrates that the proposed model is able to outperform traditional causal inference models, while simultaneously exhibit state-of-the-art performance for the estimation of average treatment effect (ATE).
%
Summarizing, the main contributions of this research are:
\begin{itemize}
	\item we propose a new causal inference model, named C-XGBoost, for the prediction of potential outcomes.
	\item we propose a new loss function for training the proposed C-XGBoost model.
	\item we provide strong empirical and statistical evidence about the prediction effectiveness and accuracy of the proposed model.
\end{itemize}

The remainder of this work is organized as follows: Section~\ref{Sec:2} provides a brief discussion of state-of-the-art models on treatment effect estimation. Section~\ref{Sec:3} presents in detail the proposed causal inference model, while Section~\ref{Sec:4} presents the used datasets in this research. Section~\ref{Sec:5} demonstrates the numerical experiments, highlighting the evaluation of causal inference models. Finally, Section~\ref{Sec:6} discusses the proposed research, summarizes its conclusions and provides some interesting ideas for future work.

\clearpage

\section{Related work}\label{Sec:2}

Causal effect estimation constitutes a complex and long-studied problem, which has recently drawn considerable attention from the machine learning community. During the last decade, a growing number of data-driven, models have been proposed for causal effect estimation, providing promising outcomes. Next, we provide a brief description of the most elegant and efficient tree-based and neural network-based causal inference models.

Kunzel et al. \cite{kunzel2019metalearners} proposed a new class of model for predicting treatment effects. The main idea is to estimate the outcome by using  all of the features together with the treatment indicator, without proving the latter any special role. In simple words, the treatment indicator is included and processed by the based learner like any other feature. R-Forest is probably the most efficient model of this class, which utilized Random-Forest \cite{breiman2001random} as base learner. However, this approach has two disadvantages; (i) in case the treatment and control groups are very different in covariates, a single model is probably not sufficient to encode the different relevant dimensions and smoothness of features for both groups \cite{alaa2018limits}; (ii) a tree-based base learner may completely ignore the treatment assignment by not choosing/splitting on it \cite{kunzel2019metalearners}.

Wager and Athey \cite{wager2018estimation} proposed Causal Forest (C-Forest), a non-parametric ensemble tree-based model, which extends the widely used R-Forest for the estimation of heterogeneous treatment effect. Specifically, C-Forest is composed by causal trees, which estimate the effect of the intervention at the trees' leaves. A considerable advantage of C-Forest is that its performance is not affected by the number of covariates such as other causal inference models and it is able to exhibit notable performance even in case where the number of covariates is relative small  \cite{wager2018estimation}.

Shi et al. \cite{shi2019adapting} proposed a novel neural network model, named Dragonnet, for estimating causal effects using observational data. The proposed model focuses on accurate estimations of both population and individual causal effects, by exploiting the sufficiency of propensity score (i.e., the probability for a sample to be assigned to the treatment group given its characteristics). Finally, the authors proposed the integration of a procedure based on non-parametric estimation theory, called targeted regularization and as experimentally proved that it is able further improve the estimation of treatment effects.

Kiriakidou and Diou \cite{kiriakidou2023integrating} developed a new methodology for causal effect estimation, named Nearest Neighboring Information for Causal Inference (NNCI). Their main objective was to improve the performance of neural network-based models for causal inference. The advantage of NNCI methodology is that it enriches the neural network model's inputs with information from neighboring outcomes in both control and treatment groups, contained in the training dataset along with the covariates. Furthermore, the authors integrated NNCI methodology on the state-of-the-art models, TARnet \cite{shalit2017estimating}, NEDnet \cite{shi2019adapting} and Dragonnet \cite{shi2019adapting}. The numerical experiments demonstrated that the adoption of the proposed methodology lead to more accurate estimations of treatment effects from the models $k$NN-Dragonnet, $k$NN-TARnet and $k$NN-NEDnet compared to their corresponding state-of-the-art models. 

In this research, we propose a new model for the prediction of potential outcomes, named C-XGBoost. The rationale behind our approach is to exploit both the superiority of tree-based models for handling tabular data and the significant property of causal inference neural network-based models to learn representations that are useful for estimating the outcome for both the treatment and non-treatment cases. Additional advantages of the proposed model are that (i) it is able to efficiently handle features with missing values requiring minimum pre-processing effort and (ii) it is equipped with regularization techniques to avoid overfitting/bias. Moreover, a new loss function was proposed for efficiently training the proposed causal inference model. The comprehensive experimental analysis provides strong empirical and statistical evidence, which secure the effectiveness of proposed approach.

\section{C-XGBoost model for treatment effect estimation}\label{Sec:3}

In this section, we provide a detailed description of the proposed tree-based C-XGBoost model for causal effect estimation. The main idea behind the proposed approach is to exploit the strong prediction abilities of eXtreme Gradient Boosting (XGBoost) algorithm together with the notable property of causal inference neural networks to learn representations that are useful for estimating the outcome for both the treatment and non-treatment cases.

It is worth mentioning that XGBoost \cite{chen2016xgboost} is a powerful ensemble tree-based algorithm, which employs a gradient boosting framework for sequentially building a series of decision trees, each one attempting to correct the errors of the previous one. This iterative process allows XGBoost to gradually refine its predictions, making it particularly effective in handling complex relationships within the data. 

A remarkable advantage of XGBoost is that it is able to build multi-output trees with multiple outputs in contrast to many traditional tree-based models. Therefore, by exploiting this ability though the proper modification of the loss function, we are able to develop a powerful tree-based ensemble model, which is able to learn representations and estimating the potential outcomes for both the treatment and non-treatment cases.

\begin{figure}[!ht]
	\centering
	\includegraphics[width=.75\textwidth]{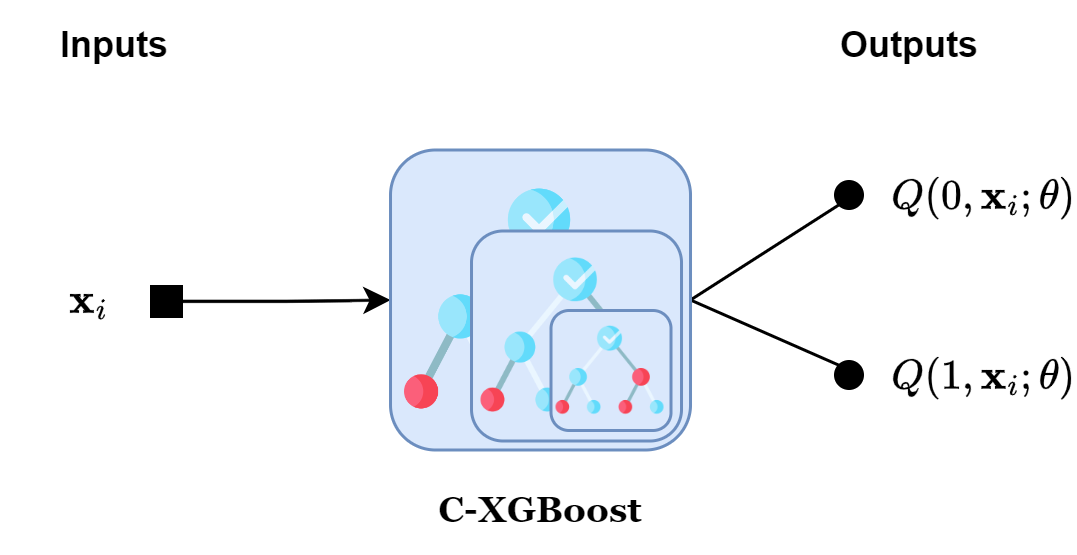}
	\caption{Proposed C-XGBoost architecture}\label{Figure:C-XGBoost}
\end{figure}

Figure \ref{Figure:C-XGBoost} presented a high-level description of C-XGBoost's architecture, which takes as input 
an instance $\mathbf{x}_i$ and calculates the prediction of the conditional outcomes $Q(0,\mathbf{x}_i;\theta)$ and 
$Q(1,\mathbf{x}_i;\theta)$ for control and treatment groups, respectively, where $\theta$ is the vector with the model's parameters. More specifically, at each iteration the C-XGBoost builds a multi-output tree with the size of leaf equals to the number of targets.\vspace{-.5cm}

For training C-XGBoost model, we define the following loss function, which constitutes a modification of mean square loss:
\begin{equation}
	\mathcal{L}(\mathbf{X};\theta) = \frac{1}{n} \sum_i \mathcal{L}_1(\mathbf{x}_i;\theta) + \mathcal{L}_2(\mathbf{x}_i;\theta),
\end{equation}
where
\begin{eqnarray}	
	\mathcal{L}_1(\mathbf{x}_i;\theta) &=& (1 - t_i)(Q(0,\mathbf{x}_i;\theta)-y_i)^2,\\
	\mathcal{L}_2(\mathbf{x}_i;\theta) &=& t_i(Q(1,\mathbf{x}_i;\theta)-y_i)^2.
\end{eqnarray}

\noindent Notice that since C-XGBoost does not provide automatic differentiation the gradient of $\mathcal{L}$ is calculated by
$$
\frac{\partial \mathcal{L}}{\partial Q(t_i,\mathbf{x}_i;\theta)} = 
\left\{
\begin{array}{ll}
	2(1-t_i)(Q(0,\mathbf{x}_i;\theta)-y_i),\quad\quad & t_i = 0;\\[.5cm]
	2t_i(Q(1,\mathbf{x}_i;\theta)-y_i)),              & t_i = 1.
\end{array}
\right.
$$
while all the elements of the Hessian matrix are equal to 2.

Finally, it is worth mentioning that some additional advantages of the proposed model, inherited by the traditional XGBoost algorithm, are that (i) it is able to efficiently handle features with missing values, which allows it to handle real-world causal inference data with missing values without requiring considerable preprocessing effort, (ii) it possesses built-in support for parallel processing, making it possible to train models on large datasets in a reasonable amount of time, (iii) it includes regularization procedures to avoid overfitting/bias.

\section{Data}\label{Sec:4}

The evaluation of robustness and effectiveness of causal inference models is generally regarded as a challenging task due to the inherent absence of the counterfactual outcomes \cite{zhou2022cycle}. In order to mitigate this issue, we employed two collections of semi-synthetic datasets:

\begin{itemize}
	\item \textbf{Synthetic} consists of a collection of toy causal-inference classification datasets which was originally introduced by Louizos et al. \cite{louizos2017causal}. Its generation is based on the hidden confounder variable $W$ using the following
	process:
	\begin{eqnarray}
		w_i &\sim& \textnormal{Bern}(0.5),\\
		t_i\, |\, w_i &\sim& \textnormal{Bern}(0.75 w_i + 0.25(1-w_i)),\\
		\mathbf{x}_i\, |\, w_i &\sim& \mathcal{N}(w_i, \sigma_{z_1}^2 w_i + \sigma_{z_0}^2 (1-w_i)),\\ 
		y_i\, |\, t_i, w_i &\sim& \textnormal{Bern}(f(3(w_i+2(2t_i-1))),
	\end{eqnarray}
	where $f$ is the logistic sigmoid function, $\sigma_{z_0} = 3$ and $\sigma_{z_1}= 5$.,
	while the proxy to $\mathbf{x}_i$ and the treatment $t_i$ consist of a mixture of Gaussian and Bernoulli distribution, respectively. In our experiments, the number of covariates was set to 1000, while each dataset contain 5000 samples as in \cite{kiriakidou2023integrating,louizos2017causal}. 
	
	\item \textbf{ACIC} consists of a collection of semi-synthetic datasets, which was developed for the 2018 Atlantic Causal Inference Conference competition data \cite{shimoni2018benchmarking}. Each dataset was received from the linked birth and infant death data \cite{macdorman1998infant} and constitutes a sample from a distinct distribution. The instances in each dataset are produced through a generating process, based in different treatment selection and outcome functions. For each setting in the data generation process, we randomly selected 5 and 11 datasets of size 5000 and 10000, respectively. Notice that all datasets in this collection were partitioned into training/testing sets based on the scheme 80/20 as in \cite{kiriakidou2023integrating}.
\end{itemize}

\section{Numerical experiments}\label{Sec:5}

In this section, we provide an extensive experimental analysis and compare the performance of the proposed C-XGBoost model with that of the state-of-the-art causal inference models: R-Forest \cite{kunzel2019metalearners} C-Forest \cite{wager2018estimation}, Dragonnet \cite{shi2019adapting} and $k$NN-Dragonnet \cite{kiriakidou2023integrating}.

All evaluated models were compared as estimators (ATE within the same dataset) and as predictors (ATE across datasets) by using the following performance metrics: absolute error in ATE
$$
\left\vert \epsilon_{ATE}\right\vert = \left\vert \frac{1}{n}\sum_{i=1}^{n}(y(\mathbf{x}_i,1) - y(\mathbf{x}_i,0)) - \frac{1}{n}\sum_{i=1}^{n}(\hat{y}(\mathbf{x}_i,1) - \hat{y}(\mathbf{x}_i,0)) \right\vert,
$$
and expected Precision in Estimation of Heterogeneous Effect
$$
\textnormal{PEHE} = \displaystyle \frac{1}{n} \sum_{i=1}^{n} \biggl( (y(\mathbf{x}_i,1) - y(\mathbf{x}_i,0)) -  (\hat{y}(\mathbf{x}_i,1) - \hat{y}(\mathbf{x}_i,0))\biggl)^{2}.
$$
Notice that these metrics constitute the most widely utilized metrics for measuring the performance of causal inference models \cite{johansson2016learning,kiriakidou2022evaluation,louizos2017causal,shalit2017estimating,shi2019adapting}.

For mitigating the influence of specific simulations on the evaluation process, we employed the performance profiles of Dolan and Mor{\'e} \cite{dolan2002benchmarking}. The scope is to provide the robustness and efficiency of each evaluated model in compact form \cite{livieris2013new}. It is worth mentioning that $x$-axis displays the performance ratio, which represents a measure of how close the performance of a causal inference model is to the optimal performance and $y$-axis shows the percentage of datasets for which a given model exhibits a certain performance ratio on the $x$-axis \cite{livieris2021smoothing}. 

Furthermore, we conducted a statistical analysis employing non-parametric Friedman Aligned-Ranks (FAR) \cite{hodges2012rank} and the post-hoc Finner \cite{finner1993monotonicity} tests for ranking the evaluated models as well as examining the existence of important differences in their performance \cite{kiriakidou2024mutual}. A comprehensive description of the evaluation process methodology as well as the employed mathematical tools can be found in \cite{kiriakidou2022evaluation}.

In the sequel, we compare the performance of: 
\begin{itemize}
	\item ``\textsc{R-Forest}'', which follows S-learner'' methodology \cite{kunzel2019metalearners} using  Random-Forest \cite{breiman2001random} as base model.
	
	\item ``\textsc{C-Forest}'', which stands for C-Forest model \cite{wager2018estimation}.
	
	\item ``\textsc{Dragonnet}'', which stands for Dragonnet model \cite{shi2019adapting}.
	
	\item ``\textsc{$k$NN-Dragonnet}'', which stands for $k$NN-Dragonnet \cite{kiriakidou2022improved,kiriakidou2023integrating}.
	
	\item ``\textsc{C-XGBoost}'', which stands for the proposed C-XGBoost model.
\end{itemize}

C-XGBoost was applied using the following hyperparameters: 
$\textsf{n\_estimator}=100, 
\textsf{max\_depth}=5, 
\textsf{reg\_lambda}=1.0, 
\textsf{tree\_method}="\textsf{hist}"$ and 
$\textsf{multi\_strategy}="\textsf{multi\_output\_tree}"$, 
while the rest parameters were set as default. The state-of-the-art causal inference models were used with their original optimized settings and hyper-parameters.

\subsection{Synthetic collection of datasets}

\begin{figure}[!ht]
	\centering
	\subfigure[$|\epsilon_\textnormal{ATE}|$]{\includegraphics[width=.5\textwidth]{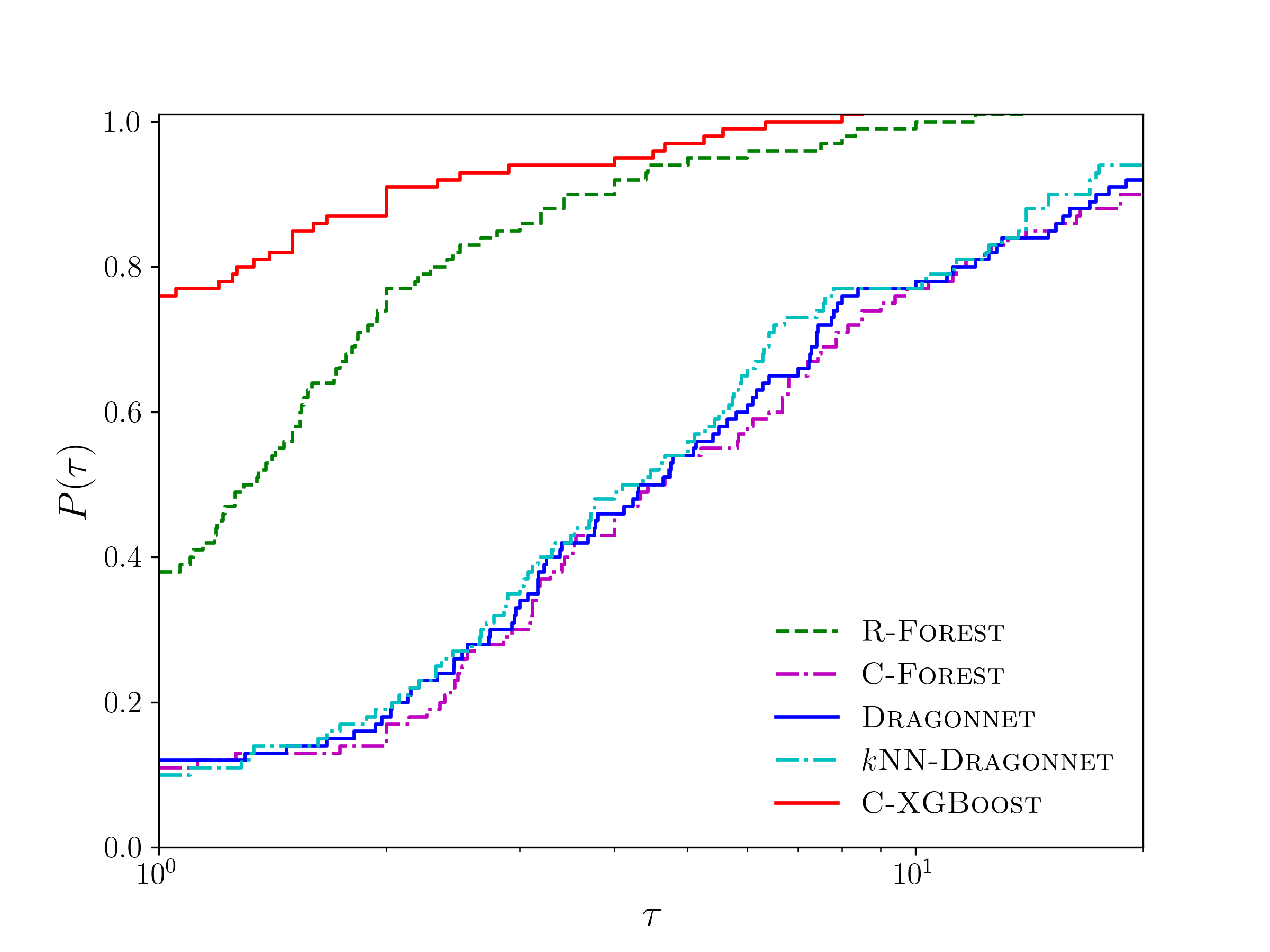}}\hfill
	\subfigure[$\epsilon_\textnormal{PEHE}$]{\includegraphics[width=.5\textwidth]{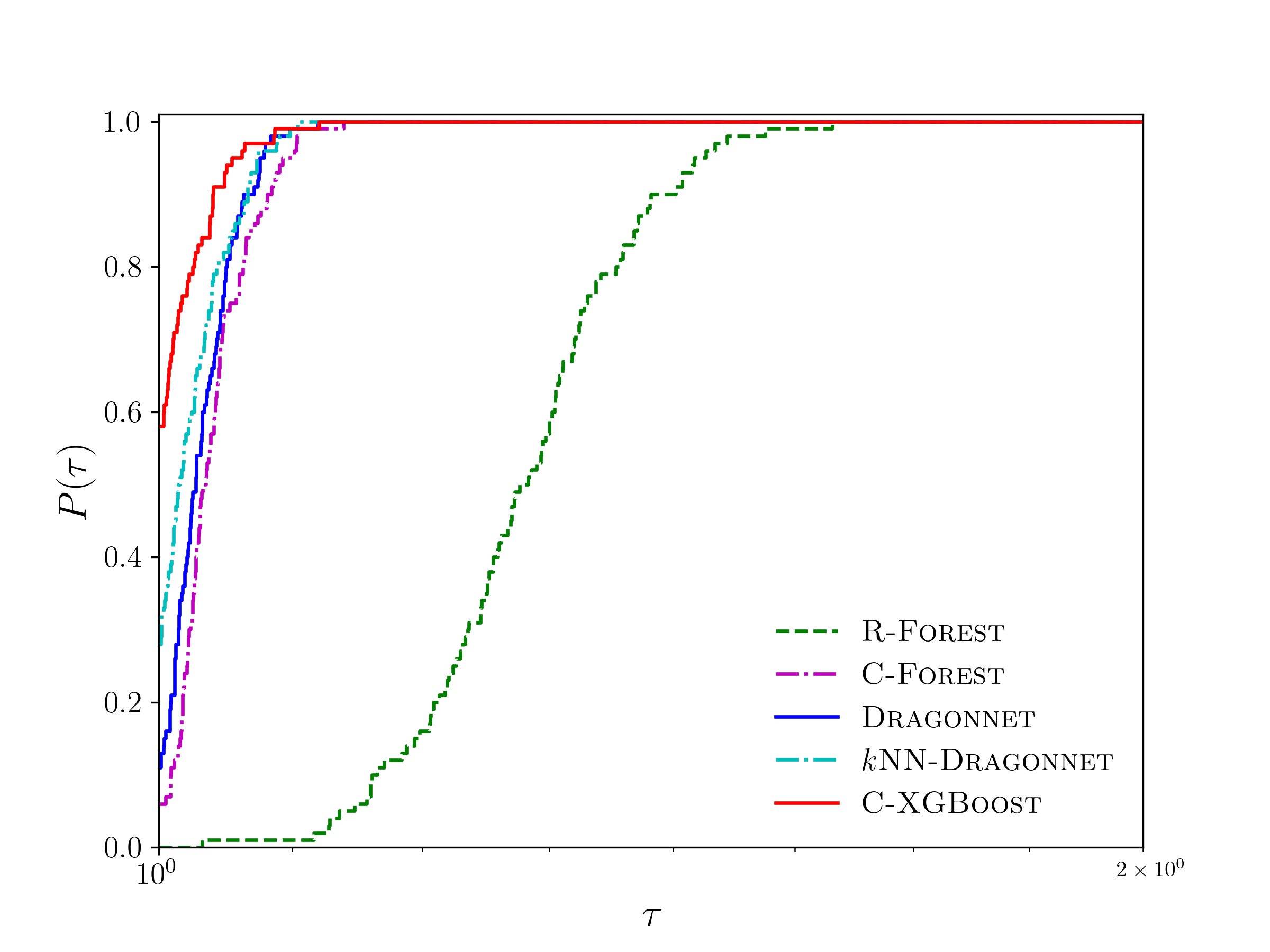}}
	\caption{$Log_{10}$ performance profiles of all evaluated models based on $|\epsilon_{ATE}|$ and PEHE for Synthetic collection of datasets}\label{Figure:Synthetic}
\end{figure}

Figure~\ref{Figure:Synthetic} presents the performance profiles for Synthetic collection of datasets as regards to both performance metrics. Clearly, the proposed \textsc{C-XGBoost} model highlights the best performance in terms of $|\epsilon_{ATE}|$ and PEHE. In more detail, \textsc{C-XGBoost} exhibits 76\% of datasets with the best (lowest)   $|\epsilon_{ATE}|$ score, followed by \textsc{R-Forest}, which presents 36\% in the same situation. The neural network-based models along with \textsc{C-Forest} perform similarly, reporting the worst performance. Relative to PEHE \textsc{C-XGBoost} presents the best performance in terms of both efficiency and robustness. \textsc{C-XGBoost} reports 58\% of datasets with the best (lowest) PEHE score, while $k$\textsc{NN-Dragonnet} reports 30\% of datasets and the rest causal inference models less than 17\%. Summarizing the performance profiles indicate that the proposed \textsc{C-XGBoost} model highlights the best overall performance as estimator as well as predictor.

Table~\ref{Table:Synthetic} summarizes the FAR and Finner post-hoc tests' results as regards both performance metrics for Synthetic collection of datasets. The Friedman statistic $F_f$ with 5 degrees of freedom are equal to 289.369 and 252.494 relative to $|\epsilon_{ATE}|$ and PEHE metrics, respectively, while the $p$-values are equal to 0 in both cases, which suggests the existence of significant differences among the performance of the causal inference models. \textsc{C-XGBoost} model reports the highest probability-based ranking, considerably outperforming all the state-of-the-art causal inference models, while Finner post-hoc test suggests the existence of statistically considerable differences among the performance of the proposed and the traditional causal inference models.

\begin{table}[!ht]
	\centering
	\setlength{\tabcolsep}{10pt}
	\renewcommand{\arraystretch}{1}
	\begin{tabular}{L{2.8cm}C{1.5cm}C{1.5cm}C{3cm}}
		\toprule
		\multirow{2}{*}{Model}   & \multirow{2}{*}{FAR} & \multicolumn{2}{c}{Finner post-hoc test} \\ \cmidrule{3-4}
		&                      & $p_F$-value &           $H_0$            \\ \midrule
		\textsc{C-XGBoost}       &        86.41         &      -      &             -              \\
		\textsc{R-Forest}        &        125.85        &  0.053579   &      Failed to reject      \\
		\textsc{$k$NN-Dragonnet} &        303.41        &  0.000000   &           Reject           \\
		\textsc{Dragonnet}       &        350.82        &  0.000000   &           Reject           \\
		\textsc{C-Forest}        &        386.02        &  0.000000   &           Reject           \\ \bottomrule
		\multicolumn{4}{c}{(a) $|\epsilon_{ATE}|$}                         \\ %
	\end{tabular}

\vspace{.3cm}
		
	\begin{tabular}{L{2.8cm}C{1.5cm}C{1.5cm}C{3cm}}
		\toprule
		\multirow{2}{*}{Model}   & \multirow{2}{*}{FAR} & \multicolumn{2}{c}{Finner post-hoc test} \\ \cmidrule{3-4}
		&                      & $p_F$-value &           $H_0$            \\ \midrule
		\textsc{C-XGBoost}       &        139.96        &      -      &             -              \\
		\textsc{$k$NN-Dragonnet} &        187.66        &  0.019582   &           Reject           \\
		\textsc{Dragonnet}       &        219.24        &  0.000139   &           Reject           \\
		\textsc{C-Forest}        &        255.28        &  0.000000   &           Reject           \\
		\textsc{R-Forest}      &          450.37        &  0.000000   &           Reject           \\ \bottomrule
		\multicolumn{4}{c}{(b) PEHE}
	\end{tabular}
	\caption{FAR test and Finner post-hoc test based on (a) $|\epsilon_{ATE}|$ and (b) PEHE for Synthetic collection of datasets}\label{Table:Synthetic}
	\end{table}\vspace{-1cm}

\subsection{ACIC collection of datasets}

\begin{figure}[!ht]
	\centering
	\subfigure[$|\epsilon_\textnormal{ATE}|$]{\includegraphics[width=.5\textwidth]{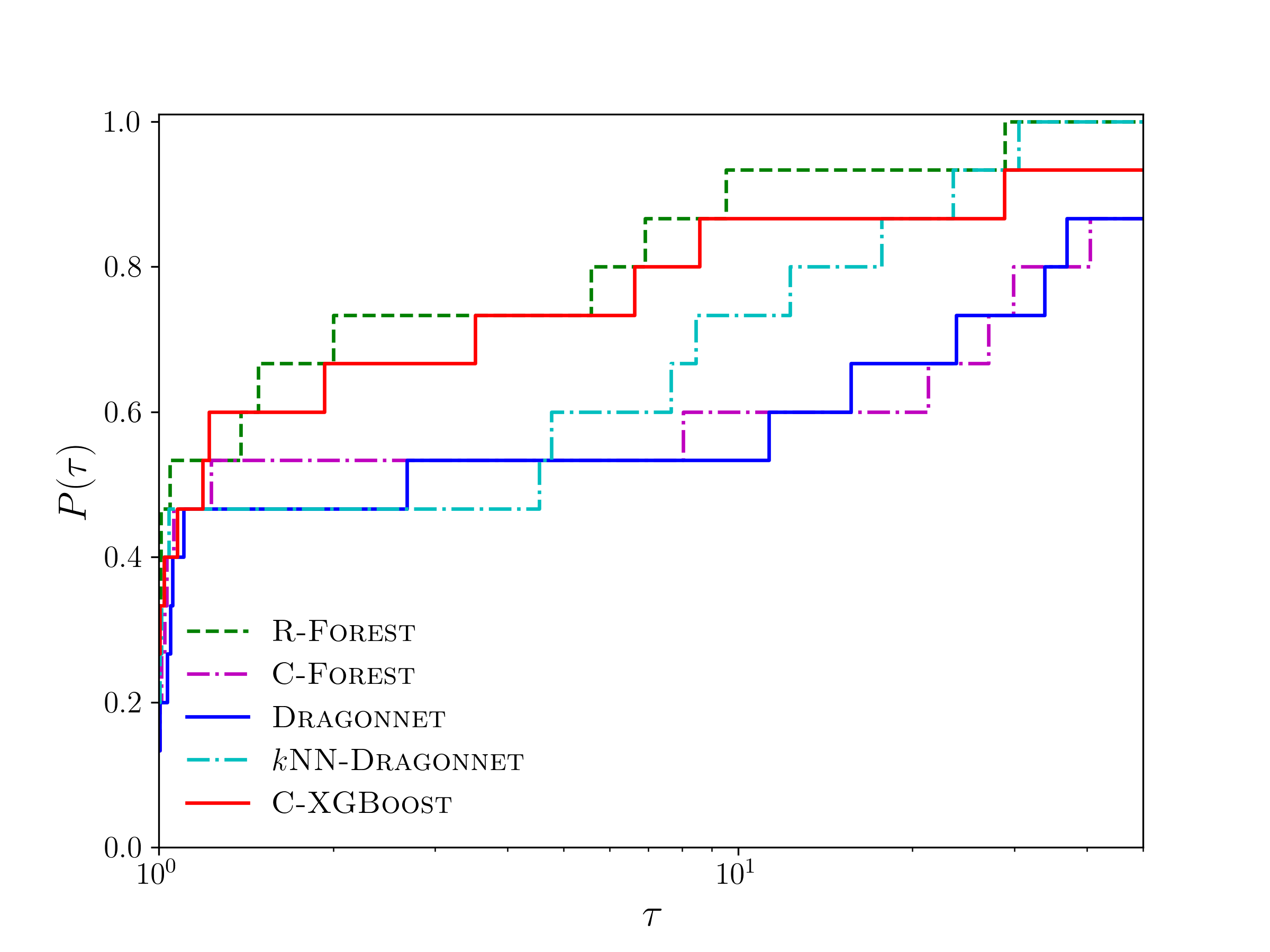}}\hfill
	\subfigure[$\epsilon_\textnormal{PEHE}$]{\includegraphics[width=.5\textwidth]{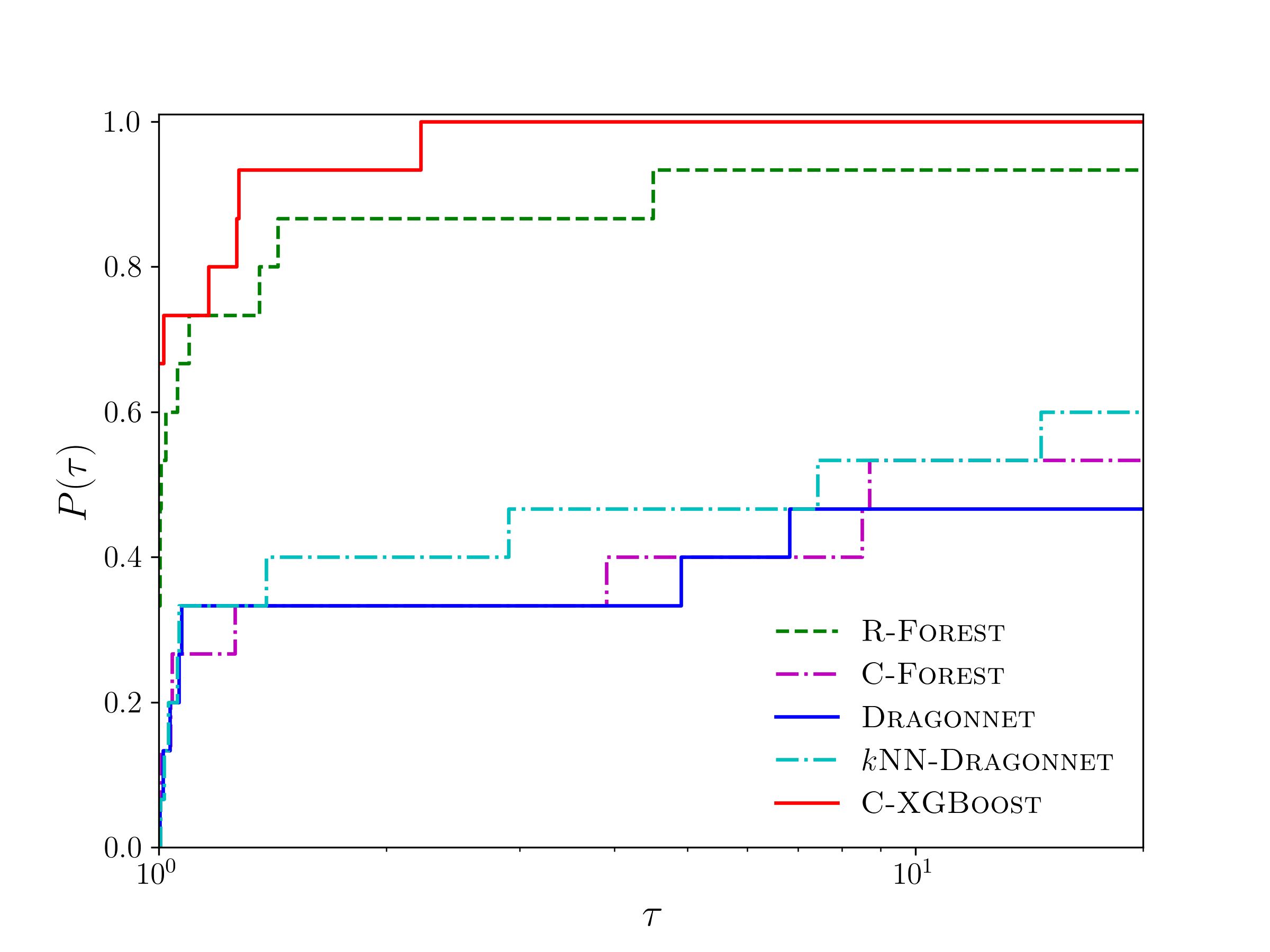}}
	\caption{$Log_{10}$ performance profiles of all evaluated models based on $|\epsilon_{ATE}|$ and PEHE for ACIC collection of datasets}\label{Figure:ACIC}
\end{figure}

Figures \ref{Figure:ACIC}(a) and \ref{Figure:ACIC}(b) present the performance profiles for ACIC collection of datasets, relative to $|\epsilon_{ATE}|$ and PEHE, respectively. The interpretation of Figure~\ref{Figure:ACIC}(a) presents that \textsc{C-XGBoost} and \textsc{R-Forest} report the best performance, exhibiting 31\% of the datasets with the lowest (best) $|\epsilon_{ATE}|$ score, followed by $k$\textsc{NN-Dragonnet}, which exhibits the best performance among the neural network based models. As regards PEHE, \textsc{C-XGBoost} present the top performance in terms of robustness, since its curve lie on the top. Furthermore, it exhibits 62.5\% of datasets with the best (lowest) PEHE score, which implies that it illustrates the best performance in terms of efficiency. It is also worth mentioning that the neural network-based models as well as \textsc{C-Forest} report the worst overall performance.

Table~\ref{Table:ACIC} presents the statistical analysis of all causal inference models for ACIC collection of datasets. The Friedman statistic $F_f$ with 5 degrees of freedom are equal to 16.76 and 24.54 relative to $|\epsilon_{ATE}|$ and PEHE metrics, respectively. The $p$-values are equal to 0.00215 and 0.00006, respectively, which suggests the hypothesis that the evaluated causal inference models perform similarly is rejected. 
In terms of $|\epsilon_{ATE}|$, the statistical analysis suggests that \textsc{R-Forest} and \textsc{C-XGBoost} report the highest probability ranking, followed by \textsc{$k$NN-Dragonnet} while Finner post-hoc test indicates that there exists no signficant differences between their performances. As regards PEHE metric, \textsc{C-XGBoost} presents the highest FAR ranking score, followed by \textsc{R-Forest} and \textsc{$k$NN-Dragonnet} models.

\begin{table}[!ht]
	\centering
	\setlength{\tabcolsep}{10pt}
	\renewcommand{\arraystretch}{1}
	\begin{tabular}{L{2.8cm}C{1.5cm}C{1.5cm}C{3cm}}
		\toprule
		\multirow{2}{*}{Model}   & \multirow{2}{*}{FAR} & \multicolumn{2}{c}{Finner post-hoc test} \\ \cmidrule{3-4}
		&                      & $p_F$-value &           $H_0$              \\ \midrule
		\textsc{R-Forest}        &        24.26         &      -      &             -                \\
		\textsc{C-XGBoost}       &        26.86         &  0.743890   &      Failed to reject        \\
		\textsc{$k$NN-Dragonnet} &        39.13         &  0.081474	  &      Failed to reject        \\
		\textsc{Dragonnet}       &        47.93         &  0.005873   &           Reject             \\
		\textsc{C-Forest}        &        51.80         &  0.002161   &           Reject             \\ \bottomrule
		\multicolumn{4}{c}{(a) $|\epsilon_{ATE}|$}
		\\
		\\
		\toprule
		\multirow{2}{*}{Model}   & \multirow{2}{*}{FAR} & \multicolumn{2}{c}{Finner post-hoc test} \\ \cmidrule{3-4}
		&                      & $p_F$-value &           $H_0$              \\ \midrule
		\textsc{C-XGBoost}       &        23.06         &      -      &             -                \\
		\textsc{R-Forest}        &        23.13         &  0.993316   &      Failed to reject        \\
		\textsc{$k$NN-Dragonnet} &        41.53         &  0.026997   &           Reject             \\
		\textsc{C-Forest}        &        49.46         &  0.001817   &           Reject             \\
		\textsc{Dragonnet}       &        52.80         &  0.000747   &           Reject             \\ \bottomrule
		\multicolumn{4}{c}{(b) PEHE}                          
	\end{tabular}
	\caption{FAR test and Finner post-hoc test based on (a) $|\epsilon_{ATE}|$ and (b) PEHE for Synthetic collection of datasets}\label{Table:ACIC}
\end{table}

\section{Discussion \& conclusions}\label{Sec:6}

The main contribution of this research is a new tree-based model, named C-XGBoost, for the prediction of potential outcomes for causal effect estimation. The key idea behind the proposed approach is to exploit the strong prediction abilities of XGBoost algorithm together with the remarkable property of causal inference neural network-based models to learn representations that are useful for estimating the outcome for both the treatment and non-treatment cases. Additional advantages of the proposed C-XGBoost model are that it is able to efficiently handle features with missing values requiring minimum pre-processing effort as well as it is equipped with regularization techniques to avoid overfitting/bias. Furthermore, a new loss function was proposed for efficiently training the proposed causal inference model.

C-XGBoost model was evaluated against state-of-the-art tree-based and neural network-based models on two semi-synthetic collections of datasets. The comprehensive experimental analysis illustrated that the proposed C-XGBoost model outperformed traditional models as an estimator and predictor. In addition, we highlighted empirical and statistical evidence about the efficiency and effectiveness of the proposed model, which was theoretical secured by the performance profiles of Dolan and Mor{\'e} \cite{dolan2002benchmarking} as well as by non-parametric Friedman Aligned-Ranks (FAR) \cite{hodges2012rank} and the post-hoc Finner \cite{finner1993monotonicity} tests.

Nevertheless, a limitation of this work can be considered the potential influence of hyperparameters settings to the efficiency of proposed model since its sensitivity to different configuration settings is unclear. Possibly, efficient hyperparameter tuning procedures may increase the effectiveness and robustness of the proposed model; hence, a comparison study using various hyperparameter settings is considered as a priority in our research tasks. It is worth mentioning that addressing this limitation through thorough experimentation not only contributes to the understanding of the C-XGBoost's behavior, but also may provide insights for future research and the development of improved causal inference models. Another limitation could be considered the fact that the evaluation process was performed on two collections of datasets. Therefore, the  application of C-XGBoost on real-world datasets is certainly included in our future work. Our primary aim is to study C-XGBoost's performance and fully capture the diverse range of scenarios and complexities present in as much as possible causal inference application domains.

Our future work is concentrated on further enhancing the robustness and predictive accuracy of the proposed C-XGBoost model by incorporating a regularization procedure for inducing bias. More specifically, an interesting idea is the employment of targeted regularization \cite{shi2019adapting}, which consists a modification to the loss function based on non-parametric estimation theory. Through experimentation and empirical evaluations, the primary aim is to gain insights into the implications of this regularization approach on the overall performance and robustness of our predictive model. Finally, another promising approach for research is the employment of an intelligent framework \cite{livieris2023advanced} for providing explainability functionalities and tools and provide insights into the decision-making process.

\subsubsection{Acknowledgements} The work leading to these results has received funding from the European Union's Horizon 2020 research and innovation programme under Grant Agreement No. 965231, project REBECCA (REsearch on BrEast Cancer induced chronic conditions supported by Causal Analysis of multi-source data).

\bibliographystyle{unsrtnat}
\bibliography{bibliography}

\end{document}